\documentclass{acm_proc_article-sp}

\usepackage[named]{algo}
\usepackage{amsmath}
\newcommand{\beq}{\begin{equation}}
\newcommand{\eeq}{\end{equation}}
\begin{document}
\title{Automatic Identification and Data Extraction
from 2-Dimensional Plots in Digital Documents}
\numberofauthors{5} 
\author{
\alignauthor
William Brouwer\\
       \affaddr{Pennsylvania State University}\\
    \email{wjb19@psu.edu}
% 2nd. author
\alignauthor
Saurabh Kataria\\
       \affaddr{Pennsylvania State University}\\
    \email{saurabh@psu.edu}
% 3rd. author
\alignauthor Sujatha Das\\
       \affaddr{Pennsylvania State University}\\
        \email{gud111@psu.edu}
       \and  % use '\and' if you need 'another row' of author names
% 4th. author
\alignauthor Prasenjit Mitra\\
       \affaddr{Pennsylvania State University}\\
        \email{pmitra@ist.psu.edu}
       % 5th. author
\alignauthor C. L. Giles\\
       \affaddr{Pennsylvania State University}\\
        \email{giles@ist.psu.edu}
}
\bibliographystyle{plain}
\maketitle

\begin{abstract} \small\baselineskip=9pt
Most search engines index the textual content of documents in
digital libraries.  However, scholarly articles frequently report
important findings in figures for visual impact and the contents of these figures are not indexed. These contents are often invaluable to the researcher in various fields, for the purposes of direct comparison with their own work. Therefore, searching for figures and extracting figure data are important problems.  To the best of our knowledge, there exists no tool to automatically extract data from figures in digital documents. If we can extract
data from these images automatically and store them in a database,
an end-user can query and combine data from multiple digital documents simultaneously and efficiently. We propose a framework 
based on image analysis and machine learning to 
extract information from 2-D plot images and store them in a database.
The proposed algorithm identifies
a 2-D plot and extracts the axis labels, legend and the data points from the 2-D plot. We also segregate overlapping shapes that correspond to different data points. We demonstrate performance of individual algorithms, using a combination of generated and real-life images.
\end{abstract}
\category{Data Mining/Extraction}{Information Systems Applications}
 \terms{Information Extraction, Machine Learning, Metadata}
\section{Introduction}

A wide variety of quantitative information is summarized and visually presented using 2-D plots, including scientific results, business performance reports, time series, etc. The embedded information is invaluable in that once extracted, the data can be indexed and
the end-user has the ability to query the data, and operate directly on the data. However, in order to extract information from figures without manual intervention, we must identify 2-D plot figures, segment the plots to extract the axes, the legend and the data sections, extract the labels of the axes, separate the data symbols from the text in the legend, identify data points and segregate overlapping data points. Performing all of these tasks automatically with high precision is a challenging problem and we believe that ours is the first attempt to achieve this goal. This paper is devoted to a subset of the overall process, specifically the identification of 2-D plots and disambiguation of overlapping data points. We perform content-based image analysis to identify appropriate features that characterize a 2-D plot from other figure types. Li, et al., \cite{lig00} have shown that the histogram distribution of the wavelet co-efficients can effectively be utilized as a global image feature for picture and non-picture classification. We adapt these methods by using additional features including line features determined after edge detection and hough transform, and the text surrounding the figure, e.g. the figure caption. Identifying data points from an image is a hard problem especially when multiple data points overlap. Typically, a figure uses common symbols (triangle, square, circle etc.) to designate a series of data points in a two dimensional space.
When data points overlap, the resulting irregular shape does not exactly match with any regularly shaped data point. To extract data precisely from figures in digital documents, one must segregate the overlapping shapes and identify the shape and the center of mass of each overlapping data
point. We employ simulated annealing, a stochastic optimization method to segregate these shapes and find the method to be fairly accurate.

\section{Related Work}
The image categorization portion of our work bears a similarity to image understanding, however, we focus on deciding whether a given image contains a 2-D plot. Li et.al. \cite{lig00}
developed wavelet transform, context sensitive algorithms to perform texture based analysis of an image, in separating camera taken pictures from non-pictures. Building on this framework, Lu et.al. \cite{lumwg06} developed an automatic categorization image system for digital library documents which categorizes the images into multiple classes within non-picture class e.g. diagram, 2-D figures, 3-D figures, diagrams and other. We find significant improvements in detecting 2-D figures by substituting certain features used in \cite{lumwg06}. \cite{1304719} presents image-processing-based techniques to extract the data represented by lines in 2-D plots. However, \cite{1304719} does not extract the data represented by data points and treats the data point shapes as noise while processing the image. Our work is complimentary in that we address the question of how to extract data represented by various shapes.

\section{Preliminary}
Our algorithm segments a 2-D figure into three regions:
1) X-axis region containing X-axis labels and numerical units, i.e., area below the horizontal axis in Fig 1., 2) Y-axis containing labels and numerical units i.e. area to the left of vertical axis in Fig 1. and, 3) plotting region, which contains legend text, data points, and lines. A 2-D figure depicts a functional distribution of the form $y_i = f_i(x)~with~conditions~w_i$ where 
Y-axis and X-axis labels contain the description for $y$ and $x$ data. The legend with textual content provides the particulars for conditions $w$, and the values for these functions are represented by the data points or the lines in the plot.

\begin{figure}[h]
\begin{center}
\scalebox{0.18}{\rotatebox{0}{\includegraphics{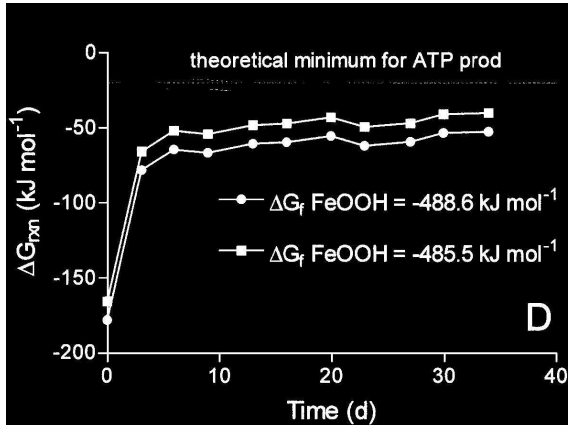}}}
\caption{A sample 2-D plot displaying experimental results reported in \protect \cite{rod02}. The areas of interest in the diagram are namely X-axis, Y-axis and plotting region.}
\end{center}
\end{figure}

\section{Method}
\subsection{Overview}
The system uses a machine-learning based classifier to identify which figures in the document are 2-D plots. An identified image is then segmented into the previously mentioned three regions. The algorithm performs connected component analysis to label each connected component in the three regions so that its shape and position can be further analyzed. Next, the candidate text components are identified based upon their mutual positioning and spacing information. This identification is based upon the intuition that the two characters appearing in the same string are very likely to be placed next to each other. Also, the spacing between them is roughly the same for any two characters appearing in any other string of text in the figure. In the next stage, we identify the data points in the plotting region. This is achieved by removing the lines from the region in a manner whereby only the data points remain; Fig. 2 depicts the entire process. 

\begin{figure}[h]

\begin{center}
\scalebox{0.35}{\rotatebox{0}{\includegraphics{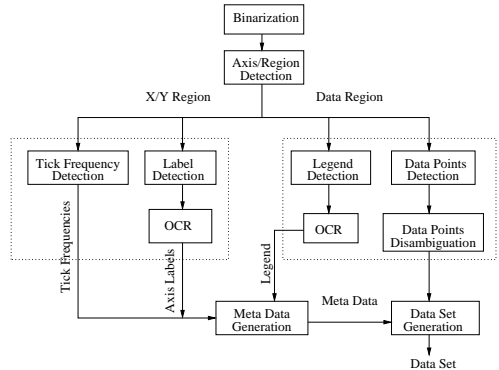}}}
\caption{Process flow of Information extraction from 2-Dimensional Plot }
\end{center}

\end{figure}

\subsection{Identification of 2-D Plots}
 
\textbf{Image segment features:}
Li, et al. \cite{lig00}, have proposed an image segmentation algorithm that divides an image into small non-overlapping blocks. They use the wavelet coefficients of each block as a localized feature to obtain  global information on the text, background and picture regions. Lu, et al.,~\cite{lumwg06} have found these localized features to be very effective in separating photo and non-photo images as well. Since 2-D plots are a subset of non-photo images, we use these features. Lu, et al.,~\cite{lumwg06} have noted that the finer aspects of colors and shades do not contribute heavily towards identifying the "semantic type" of the figure. Therefore, in extracting the image segment features, we converted each image to grayscale (portable gray map or PGM) format.

\textbf{Axes Features:}
2-D figures range from curve-fitted plots to histograms and pie-charts. We are primarily interested in 2-D plots that graph the variation of a variable with respect to another variable and the presence of co-ordinate axes is certainly a distinguishing feature of such plots. We apply the 
Hough transform \cite{hough} on the binarized image to obtain the positional information of the longest straight lines, including their mutual angles (eg., X-Y axes are othogonal) and use these as features.

\textbf{Text Features:}
From our observations, we found that authors tend to employ certain terms in writing captions for 2-D plots that are used less frequently in captions for other types of figures. For instance, re-occurring sets of words include \textit{distribution, slope, axes, plot, range}, etc. We use these words to form boolean features while training our classifier. 

Support Vector Machines(SVM) \cite{burges98} are increasingly used in both 2-class and multi-class classifications for their robustness and computational efficiency when compared to other machine learning techniques. We train our classifier based on the afore-mentioned features using an SVM, and found that a linear kernel along with the C-parameter set to $1.0$
was best suited to our purposes.

\subsection{Data Point Disambiguation}

Overlapping data points occur frequently in 2-D
plots and identifying each individual data point and its coordinates is a difficult task. We apply simulated annealing (SA) in order to resolve individual data points within a region of overlap. SA is a stochastic method, based on the Metropolis algorithm, often used in non-convex optimization problems. It bears close similarity to annealing (i.e. slow cooling) in metallurgical processes. By analogy to its physical counterpart, the optimal configuration (lowest energy $E_{min}$) is approached while the temperature $T$ is lowered. In accordance with the Metropolis algorithm, occasionally higher energy configurations $E_f > E_i$ are assumed with probability $e^{-(E_f-E_i)/T}$. The specific details of the algorithm are presented below.

We generate an 'initial configuration' image, which consists of large numbers of randomly selected candidate shapes, with random positions. Candidates are previously identified shapes extracted from the 2-D plot, using standard shape detection methods~\cite{331956}. The target image consists of overlapping data points, extracted from within the plotting region, which has failed to be classified as a particular shape. Hence, we consider two matrices with binary (boolean) values: the generated image and the original overlapping data point image. A Grammian matrix is constructed from the difference between these two matrices, the trace of which is used as a cost function, and is minimized iteratively as follows. To begin with, the coordinates of the candidate shapes are given random fluctuations, within the image boundary, which is determined by the size of the target image. In addition, point types are swapped, much like optimization within combinatorial problems such as TSP. Finally, the Euclidean distance between the centroids of identical shapes is used as a measure for removal of identical types which overlap. In this manner, the numbers, variety and coordinates of individual data points are ascertained. Carnevali, et al.,~\cite{car85}, applied simulated annealing to construct an image from known sets of shapes in the presence of noise. However, to the best of our knowledge, application of simulated annealing to disambiguate overlapping shapes is a novel contribution. 

\begin{scriptsize}
\begin{algorithm} {Data Point Disambiguation}
\qinput{ $N$ Binarized shapes, $shape[1..N]$; 
 Binarized pixel region $B$ of overlapping points, height $h$ and width $w$}
\qoutput{ Coordinates \& numbers of independent data points} \\
\textbf{ for} point-type $shape[k]$\\
$bound[k][m,n]$ = $[h-height(shape[k]), w-width(shape[k])]$ \qcom{Determine bounds $m,n$ for individual data point centroids from target image size.}\\
\qcom{ Initial centroid  for point-type $shape[k]$} \\
$centroid_i[k][i,j] = $rand$*bound$ \\
$weight[i]=1$ \qcom{ All initial weights =1}\\
$E_i = T = Cost(B,shape[k],centroid_i,weight)$ \qcom{ Initial energy  and temperature }\\
\qrepeat \qcom{rand fluctuation  to $k$th co-ordinates}\\
$centroid_f[k][i,j] += $round$($rand$*2-1)$\\
$E_f = Cost(B,shape[k],centroid_f,weight)$ \qcom{update cost after move; }\\
\qif $E_i > E_f$\\
\qthen $centroid_i=centroid_f$ \qcom{accept move}\\
\qelse accept with probability $\exp[-(E_f-E_i)/T]$ \qfi\\
\qif $\exp[-(E_f-E_i)/T] <$rand\\
\qthen $centroid_i=centroid_f$ \qcom{accept move} \qfi
\quntil $E_f < \epsilon$\\
\qif $distance(centroid_i[k][i,j],centroid_i[l][i,j]) \approx 0$ \\
\qthen $weight[k]=0$ \qfi \qcom{ Every $\alpha$ steps, remove one of two identical overlapping points $k$}\\
$T=T*(1-e)$  \qcom{Every $\beta$ steps, reduce temperature }\\
$tmp=centroid_i[k][i,j]$\\
$centroid_i[k][i,j]=centroid_i[l][i,j]$ \\
$centroid_i[l][i,j]=tmp$
\qcom{Every $\gamma$ steps, swap two point types}

\end{algorithm}  

\begin{algorithm} {Cost Calculation}
\qinput{$B, shape[k], centroid_i, weight$}
\qoutput{cost}

$C$=zeros(size($B$)) \qcom{Create empty matrix $C$ with dims. of $B$}\\

$p$=length$(weight)$\\
\qfor $k \qlet p$\\

\qdo $C[centroid_i[k][i]:X,centroid_i[k][j]:Y]~ |$ \\ $(shape[k]*weight[k])$ \qrof
\qcom{logical OR between range of indices in matrix C and candidate points of size $X,Y$ }\\

\qreturn $Trace[(B-C)'*(B-C)]$
\qcom{trace of Grammian, transpose of (B-C) times (B-C)}
\end{algorithm}

\end{scriptsize}
\section{Experiments}

In this section, we report the results obtained by
evaluating the new features for 2-D plot identification and data point disambiguation algorithms. The data set that we used for our experiments is randomly selected publications crawled from the web site of Royal Society of Chemistry {www.rsc.org} and randomly selected computer science publications from the CiteSeer digital library ~\cite{276685}  for scientific publications.

\subsection{2-D figure Classification}
 
For our classification experiments, we extracted the images from the afore-mentioned documents and had them manually tagged by two volunteers as 2-D or non 2-D. Our set consists of 2494 images, out of which 734 images are 2-D plots. As mentioned previously, we train a linear SVM(with $C=1.0$) on this dataset.

\begin{table}[h]
\begin{tabular}{|l|c|}
\hline
 Features & \% CV(\#3) accuracy \\
\hline
Only IS & 85.24\\
\hline
Only CT & 78.3    \\
\hline
IS + CA & 85.85\\
\hline
CT + CA & 80.67 \\
\hline
IS + CT & 85.85\\
\hline
All & 88.25 \\
\hline
\end{tabular}
\label{table:cvaccuracies}
\caption{Cross-validation accuracies}
\end{table}
%\end{center}

\begin{table}[hc]
\begin{tabular}{|l|c|c|}
\hline
Class & Non 2-D & 2-D \\
\hline

Non 2-D & 1393 & 67 \\
\hline
2-D & 82 & 452 \\
\hline
\end{tabular}
\label{table:cmatrixtrain}
\caption{Confusion matrix(train set)}

\end{table}

\begin{table}[hc]
\begin{tabular}{|l|c|c|}
\hline
Class & Non 2-D & 2-D \\
\hline
Non 2-D & 273 & 27 \\
\hline
2-D & 66 & 134 \\
\hline
\end{tabular}
\label{table:cmatrixtest}
\caption{Confusion matrix(sample test set)}

\end{table}

\subsubsection{Feature extraction}

Table 1 shows the 3-fold cross-validation accuracies with different
combinations of features. We use the following abbreviations: IS for image segmentation, CT for caption text, CA
for the coordinate axes. The confusion matrix over a sample test set is shown in Table 3.
For comparison purposes, we have also shown the confusion matrix over the training set in Table 2.
The libSVM software was used for support vector classification \cite{CC01a}.

\subsection{Data Point Disambiguation}
For the purposes of our experiment, 90 $\times$ 90 sized images of overlapping points were generated randomly using two types, a diamond (A) and triangle (B). Fig. 3 gives typical examples of pixel regions containing overlapping data points and the corresponding machine-learnt version; table 4 details the experimental parameters and results corresponding to fig. 3.

\begin{center}
\begin{figure}[ht]

\begin{math}
\begin{array}{cc}
\scalebox{0.8}{\rotatebox{0}{\includegraphics{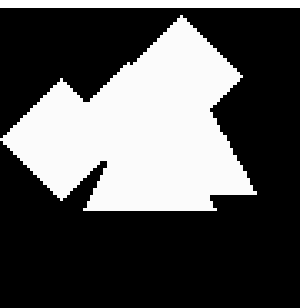}}} & \scalebox{0.8}{\rotatebox{0}{\includegraphics{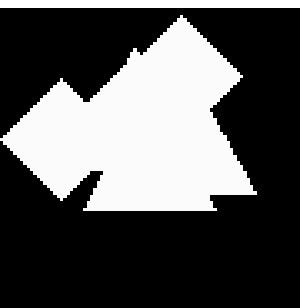}}} \\
\scalebox{0.8}{\rotatebox{0}{\includegraphics{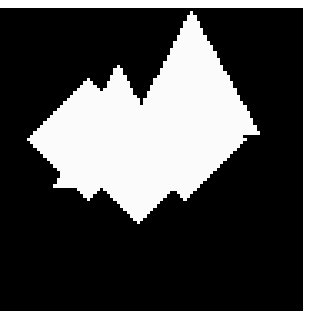}}} & \scalebox{0.8}{\rotatebox{0}{\includegraphics{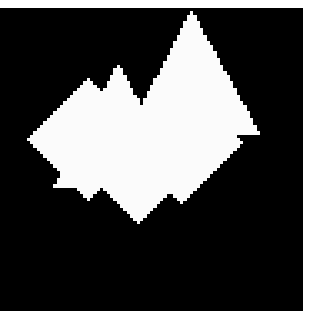}}} \\
\end{array}
\end{math}
\caption{Examples of overlapping data points (left) and machine learnt versions (right)}
\end{figure}
\end{center}

\begin{table}[!ht]
\begin{tabular}{|l|c|c|c|c|}
\hline
Iterations & Temp.  & Type & Offset & Offset\\
\ &const. & &(orig.)&(calc.) \\
\hline
10k & 0.4 & A & (11,39) & (11,40) \\
& & & (35,19) & (34,20) \\
& & & (19,4) & (20,3) \\
& & B & (21,35) & (22,35) \\
& & & (10,18) & (10,17) \\
\hline
10k & 0.3 & A & (29,24) & (29,23) \\
& & & (22,9) & (21,9) \\
& & & (23,37) & (21,39) \\
& & B & (2,39) & (2,39) \\
& & & (18,17) & (18,16) \\
\hline

\end{tabular}
\caption{Example parameters for simulated annealing
 applied to the data point disambiguation problem.}
\end{table}

Table 5 gives the overall results of these experiments using an annealing constant of 0.4 and 10k iterations. As the annealing schedule is slowed and iterations increased, the recall approaches 100\%.  A slower annealing schedule than that used here and more iterations are required as the pixel region and number of possible different data points increases. However the results are promising in that data that would traditionally be considered lost is recovered with fairly high accuracy.

\begin{table}[ht]
\begin{tabular}{|l|c|c|c|}
\hline
Shape & Total & \# Correct & \% Recall\\ \hline
Diamond & 72 & 64 & 88.9 \\
Triangle & 78 & 71 & 91.0 \\
\hline
\end{tabular}
\caption{Experimental Results for
Data-Point Disambiguation} \centering

\end{table}
\[ \]
\[ \]

\section{Conclusions and Further Work}
We have outlined a system that can identify 2-D plots in digital documents and extract data from the identified documents. Overlapping data points present a major challenge in reconstructing data series from within the plotting region, once lines are filtered from 2-D plots. We present an unsupervised machine-learning algorithm to segregate overlapping data points and identify their exact shape and location. The work presented here is currently being integrated into the overall figure extraction system. In addition, attention is being given to improving the quality of extracted textual information, to assist in indexing of figures.

%\begin{scriptsize}
\bibliography{ref.bib}
%\end{scriptsize}
\end{document}